\documentclass{article} 
\usepackage{iclr2026_conference,times}

\usepackage{hyperref}
\usepackage{url}
\usepackage{amsmath}
\usepackage{amssymb}
\usepackage{bm}
\usepackage{mathtools}
\usepackage{caption}
\usepackage{booktabs}
\usepackage[ruled, vlined]{algorithm2e}
\usepackage{wrapfig}
\usepackage{placeins}

\usepackage{listings}
\usepackage{xcolor}
\usepackage{tcolorbox}
\tcbuselibrary{listings, breakable, skins}

\newtcblisting{prompt}{
    listing only,
    breakable,
    enhanced,
    colback=gray!10,
    colframe=gray!50,
    left=5pt,
    right=5pt,
    listing options={
        basicstyle=\ttfamily\small,
        breaklines=true,
        breakatwhitespace=true
    }
}
\newcommand{\setfnt}{\mathsf}
\newcommand{\vecfnt}{\bm}

\newcommand{\R}{\mathbb{R}}
\newcommand{\Z}{\mathbb{Z}}

\newcommand{\defeq}{\coloneq}

\DeclareMathOperator{\expec}{\mathbb{E}}

\DeclareMathOperator{\softmax}{Softmax}
\DeclareMathOperator{\categorical}{Cat}

\DeclareMathOperator{\argmax}{argmax}

\DeclareMathOperator{\normal}{Normal}
\DeclareMathOperator{\uniform}{Uniform}

\title{Learning to Query History: Nonstationary Classification via Learned Retrieval}

\newcommand{\eqcont}{\textsuperscript{*}}
\newcommand{\jimaffil}{\textsuperscript{$\alpha$}}
\newcommand{\amzaffil}{\textsuperscript{$\beta$}}
\newcommand{\gtaffil}{\textsuperscript{$\gamma$}}
\author{\makebox[\textwidth][c]{\parbox{0.8\textwidth}{\centering
    Jimmy Gammell\eqcont\jimaffil{} \hspace{3em} Bishal Thapaliya\eqcont\amzaffil{} \hspace{3em} Yoon Jung\amzaffil{} \\[1em] Riyasat Ohib\gtaffil{} \hspace{3em} Bilel Fehri\amzaffil{} \hspace{3em} Deepayan Chakrabarti\amzaffil{}
}}}

\iclrfinalcopy

\track{Research}

\begin{document}

\maketitle

\renewcommand{\thefootnote}{\fnsymbol{footnote}}
\setcounter{footnote}{0}

\footnotetext[1]{
    Equal contribution.\, \jimaffil{} Purdue University, Elmore Family School of Electrical and Computer Engineering. Work done during Amazon internship.\, \amzaffil{} Amazon.\, \gtaffil{} Georgia Institute of Technology. \, Correspondence to Jim Gammell: \texttt{jgammell@purdue.edu},\, Bishal Thapaliya: \texttt{bishath@amazon.com}.
}

\begin{abstract}
   Nonstationarity is ubiquitous in practical classification settings, leading deployed models to perform poorly even when they generalize well to holdout sets available at training time. We address this by reframing nonstationary classification as time series prediction: rather than predicting from the current input alone, we condition the classifier on a sequence of historical labeled examples that extends beyond the training cutoff. To scale to large sequences, we introduce a learned discrete retrieval mechanism that samples relevant historical examples via input-dependent queries, trained end-to-end with the classifier using a score-based gradient estimator. This enables the full corpus of historical data to remain on an arbitrary filesystem during training and deployment. Experiments on synthetic benchmarks and Amazon Reviews '23 (electronics category) show improved robustness to distribution shift compared to standard classifiers, with VRAM scaling predictably as the length of the historical data sequence increases.
\end{abstract}

\section{Introduction}

Nonstationarity is a persistent problem in practical classification settings such as policy violation detection, fraud classification, and compliance, where violation patterns evolve as trends change and bad actors adapt. We consider nonstationary classification settings where $X_t, Y_t \sim p_t$ and aim to predict $Y_t$ from $X_t.$ Models train on early data ($t < t_{\mathrm{cutoff}}$) and must generalize to later times.

Deployed systems often accumulate a growing corpus of labeled historical examples that extends beyond the training cutoff up to the current time. We propose to leverage this by reframing nonstationary classification as time-series classification: rather than predicting $Y_{t_0}$ solely from $X_{t_0},$ we also feed the classifier a long sequence of historical examples: $(X_{t_{-1}}, Y_{t_{-1}}, \dots, X_{t_{-K}}, Y_{t_{-K}}).$ This allows the model to adapt to distribution shift without re-training, by observing the relationship between inputs and labels for relevant recent examples.

Such sequences may be so long that it is impractical to naively feed them as an auxiliary input to the classifier, or even to store them in accelerator memory. Additionally, the number of examples which are relevant to the current query may be small. We thus propose a system with a learned retrieval mechanism which $X_{t_0}$-conditionally samples a small number of relevant items from the full sequence, then feeds these to a downstream classifier which predicts $Y_t.$ The full system is trained end to end to optimize the classification objective, and the sequence may remain on an arbitrary filesystem during both training and deployment, with only the selected items and a small per-item key stored in accelerator memory.

To our knowledge, our work is the first to explore nonstationary retrieval-augmented classification by leveraging a large and growing corpus of historical data. We validate this approach on controlled synthetic benchmarks and a variant of the electronics category of Amazon Reviews '23 \citep{hou2024bridging}. Our experiments demonstrate that 1) the joint training procedure successfully learns to retrieve and exploit relevant historical context, 2) this improves robustness to distribution shift compared to standard classifiers, and 3) we can use corpora which are too large to fit in available accelerator memory.

\paragraph{Related work}

Nonstationary classification has previously been addressed through continual learning, which incrementally updates model weights as new data arrives \citep{kirkpatrick2017overcoming,wang2024comprehensive}, and test-time adaptation, which adapts models during inference \citep{liang2025comprehensive} using a limited sample of data from the test distribution. Our approach is complementary: we train the model to condition on a large quantity of recent context which may be from the training or test distribution, without the cost or risk of forgetting associated with weight updates.

We build on prior work in retrieval-augmented learning \citep{lewis2020rag,guu2020realm}, which has proven effective for classification under distribution shift \citep{xu-etal-2023-retrieval, long-etal-2023-adapt, lee2025ra, Fan_2025_ICCV} and weight update-free continual learning for LLMs \citep{gutierrez2025hipporag}. We extend retrieval-augmented learning to nonstationary classification settings where the corpus is a growing body of labeled historical examples which are too numerous to fit in accelerator memory.

\section{Method}\label{sec:method}

\begin{figure}[t]
   \centering
   \includegraphics{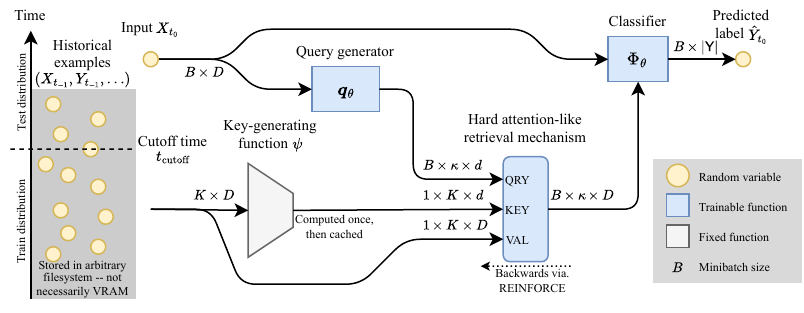}
   \caption{A diagram illustrating the architecture of our system. We learn input-dependent queries, which are used to sample from a massive corpus of historical data in a manner similar to stochastic hard attention. The input and selected corpus items are then fed to a downstream classifier which predicts the label. Through hard sampling at training time and attention masking, our implementation allows full corpus to stay on an arbitrary filesystem at both training and inference time, facilitating scaling to huge corpora.}
   \label{fig:full-system}
\end{figure}

See Fig.~\ref{fig:full-system} for a diagram of our system, and Algorithm~\ref{alg:training-step} for pseudo-code of a training step. We consider a nonstationary supervised classification setting with time $t$-indexed inputs $X_t \in \R^D$ and labels $Y_t \in \setfnt{Y}$ where $\setfnt{Y}$ is a finite set. These are jointly-distributed according to a time-dependent distribution: $X_t, Y_t \sim p_t.$ Given $X_{t_0}, Y_{t_0} \sim p_{t_0},$ we aim to train a classifier which can predict $Y_{t_0}$ from $X_{t_0}$ and a series $(X_{t_{-1}}, Y_{t_{-1}}, \dots, X_{t_{-K}}, Y_{t_{-K}})$ where $t_{-K} \leq t_{-K+1} \leq \cdots < t_0.$ Our model is trained solely on data where $t_0 < t_{\mathrm{cutoff}},$ then deployed on data where $t_0 \geq t_{\mathrm{cutoff}}.$

We associate each historical datapoint with a low-dimensional `key': $k_{t_{-m}} \defeq \psi(X_{t_{-m}}, Y_{t_{-m}})$ where $\psi: \R^D \times \setfnt{Y} \to \R^d$ is a frozen function. For example, when the inputs are text data, we may generate keys with a text embedding model \citep{zhang2025qwen3}. We assume that $d \ll D$ is sufficiently small that we can store all $K$ keys from our historical example sequence in VRAM. For large $K$ there may be a significant up-front cost to computing these keys. However, because $\psi$ is frozen, each key can be computed once and subsequently cached and reused.

Our system consists of several modules, each parameterized by weights $\vecfnt{\theta}.$ A `query generator' $\vecfnt{q}_\theta: \R^D \to \R^{\kappa \times d}$ generates $\kappa$ input-dependent queries: $(q_1, \dots, q_\kappa) = \vecfnt{q}_\theta(X_{t_0}).$ For each query we compute logits over the set of historical example indices $\{1, \dots, K\}$ as $u_m \defeq (q_m^\top k_n / \sqrt{d}: n=1, \dots, K)$ (as in an attention mechanism). We then sample $\kappa$ timesteps without replacement by iteratively sampling $\hat{t}_m \sim \operatorname{Categorical}(\softmax(u_m)),$ then setting the logit for index $\hat{t}_m$ to $-\infty$ in all subsequent queries. After sampling these timesteps, we load the historical examples $(X_{\hat{t}_1}, Y_{\hat{t}_1}, \dots, X_{\hat{t}_\kappa}, Y_{\hat{t}_\kappa})$ into accelerator memory.

Finally, we predict the label with a classifier $\Phi_\theta: \R^D \times (\R^D \times \setfnt{Y})^\kappa \to \Delta^{\lvert\setfnt{Y}\rvert-1}.$ We train the full system end-to-end to minimize the modeled negative log-likelihood of the data:
\begin{equation}\label{eqn:objective}
   \min_{\vecfnt{\theta}} \quad \mathcal{L}(\vecfnt{\theta}) \defeq -\expec \log \Phi_\theta\big( Y_{t_0} \mid X_{t_0}, X_{\hat{t}_1}, Y_{\hat{t}_1}, \dots, X_{\hat{t}_\kappa}, Y_{\hat{t}_\kappa} \big)
\end{equation}
where $\expec$ is taken over the data distribution and the randomness of the selection mechanism.

\paragraph{Backpropagation through the discrete retrieval mechanism} In order to solve Eqn.~\ref{eqn:objective} with stochastic gradient-based optimizers, we must backpropagate through a discrete categorical distribution. We do so using a score-based gradient estimator \citep{williams1992reinforce} with a greedy baseline \citep{rennie2017self}. This is conceptually similar to the truncated marginalization approximation of \citet{lewis2020rag,guu2020realm}. Straight-through \citep{bengio2013estimating} and relaxation \citep{jang2017categorical}-based gradient estimators are more popular for use cases of this nature. However, we cannot use them here because they entail approximating the `hard' discrete distribution with a `soft' continuous one during training, and would thus require storing all $K$ historical examples in accelerator memory.

\paragraph{Memory footprint} We attain significant memory savings by storing the broader corpus of historical examples off-device. We can render the memory footprint of the keys independent from minibatch size by storing a single copy of all $\tilde{K} \in \Theta(K)$ keys in our dataset in VRAM, and setting elements of our retrieval mechanism's logits $u_m$ to $-\infty$ where they correspond to keys not in $k_{\hat{t}_1}, \dots, k_{\hat{t}_K}.$ We can further reduce the VRAM consumption due to the logits over the corpus from $O(\kappa K)$ to $O(K)$ by computing them one at a time and freeing them after sampling corpus indices. When using a minibatch size $B,$ this leads to a memory complexity of
\begin{equation}\label{eqn:vram-scaling}
   O\Big(\underbrace{dK}_{\text{keys}} + \underbrace{BK}_{\text{logits over corpus}} + \underbrace{B\kappa D}_{\text{retrieved items}} + \underbrace{f(B, \kappa, d, D)}_{\text{model weights + activations}}\Big).
\end{equation}

\section{Experiments}

\begin{figure}[t]
   \centering
   \includegraphics{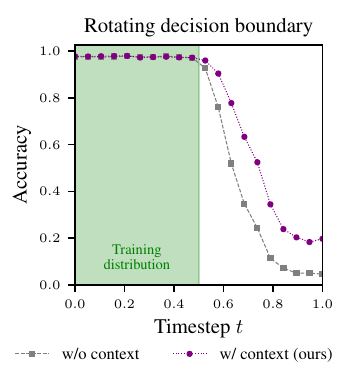} \hfill
   \includegraphics{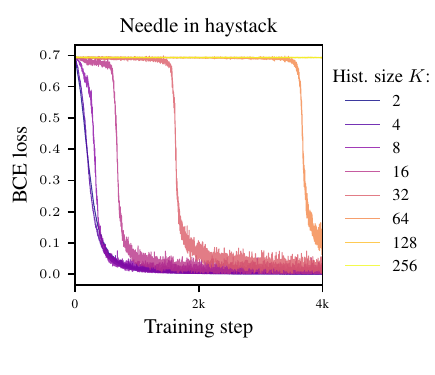}
   \caption{Synthetic settings. (\textbf{left}) Accuracy vs. time in a nonstationary binary classification setting with rotating decision boundary. Models train on $t \in [0, 0.5]$ and test on $t \in [0, 1].$ Our method leverages historical context to mitigate performance degradation beyond the training distribution. (\textbf{right}) Training dynamics on a `needle in haystack' task where the label is encoded in one historical item. The system learns to retrieve the correct item after a random exploration phase with duration scaling linearly with the number of items.}
   \label{fig:toy-settings}
\end{figure}

We run several experiments to verify that our system is scalable and improves robustness over standard classifiers in nonstationary settings. We provide a high-level summary here and defer details to Appendix~\ref{sec:appendix-experiments}.

\paragraph{Controlled synthetic settings}

We validate our system in two synthetic settings (Fig.~\ref{fig:toy-settings}). The first `rotating decision boundary' setting is a binary classification problem where the optimal decision boundary smoothly rotates $\pi$ radians as time varies from $t=0$ to $1.$ Our system mitigates performance degradation due to this distribution shift by leveraging historical context. In the second `needle in haystack' setting, the label is encoded in a single historical example, and our system cannot surpass random performance unless it learns to retrieve this example. Our system learns to identify and retrieve the correct example, demonstrating that we can successfully jointly train the retrieval mechanism and classifier using only the classification objective.

\paragraph{Sentiment classification on Amazon Electronics Reviews}

\begin{figure}[t]
   \centering
   \includegraphics{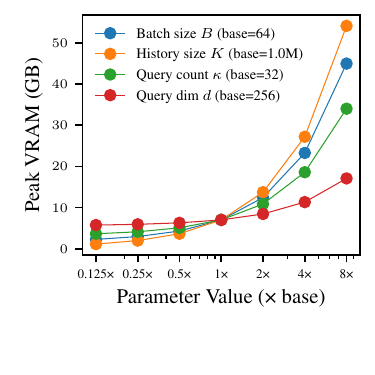}
   \hfill
   \includegraphics{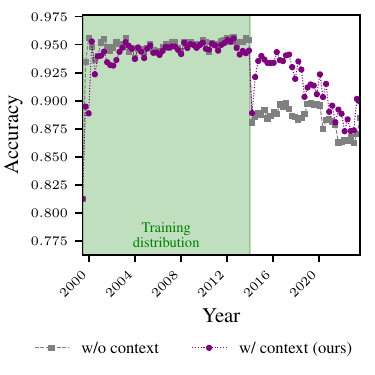}
   \caption{Amazon Electronics Reviews. (\textbf{left}) Peak VRAM consumption during training vs. hyperparameters. Scaling is consistent with Eqn.~\ref{eqn:vram-scaling}. (\textbf{right}) Accuracy over time for models trained on pre-2014 data. Our method leverages historical context to mitigate performance degradation on out-of-distribution data beyond 2014.}
   \label{fig:amazon-reviews-figures}
\end{figure}

We evaluate scalability using subsets of the electronics category of Amazon Reviews '23 \citep{hou2024bridging} (43.9M total reviews spanning 1996--2023), training on pre-2014 data. We do sentiment classification where the classification threshold changes over time (ratings $>3$ positive pre-2014, $>1$ positive post-2014), creating substantial distribution shift. Text features are embedded using Qwen3-Embedding-8B \citep{zhang2025qwen3}, with truncated versions of these used as keys. Our classifier and query generator are implemented with Perceiver blocks \citep{jaegle2021perceiver} and trained for 100 epochs with batch size 256 using AdamW. Fig.~\ref{fig:amazon-reviews-figures} illustrates that peak VRAM during training scales affinely with the salient parameters of Eqn.~\ref{eqn:vram-scaling}, and that our system successfully exploits historical labels to mitigate performance degradation on the out-of-distribution post-2014 data. In Fig.~\ref{fig:amazon-reviews-figures} (right) we configure our system to retrieve only historical labels $Y_{t_{-m}}$ rather than full examples $(X_{t_{-m}}, Y_{t_{-m}}).$ While our framework supports both, we find that in this setting label-only retrieval improves robustness whereas full-feature retrieval does not (see Appendix \ref{sec:label-vs-nolabel}).

\section{Limitations and future directions}

Our work motivates several future directions for retrieval-augmented nonstationary classification: First, our system requires careful tuning to manage the high variance of the score-based gradient estimator and to prevent the classifier from ignoring historical context early in training. Exploring alternate gradient estimators, variance-reducing baselines, and regularization strategies could improve training stability. Second, our assumption of a frozen key-generating function limits retrieval expressiveness; periodic key updates would likely enhance performance. Third, while our approach improves robustness, performance still degrades on out-of-distribution data. It could complement continual learning methods by reducing the minimum weight-update frequency to maintain performance. Finally, retrieval from large corpora unavoidably incurs computational and memory overhead relative to standard classifiers, which must be traded off with robustness improvements.

\bibliography{iclr2026_conference}
\bibliographystyle{iclr2026_conference}

\appendix

\section{Algorithm Pseudocode}

\begin{algorithm}
   \caption{Pseudocode for a single training step.}
   \label{alg:training-step}
   \SetKwInOut{KwIn}{Input}
\SetKwInOut{KwOut}{Output}
\KwIn{Initial weights $\theta_{\mathrm{in}} \in \setfnt{\Theta},$ input $x \in \setfnt{X},$ label $y \in \setfnt{Y},$ corpus $\vecfnt{c} \equiv (c^{(1)}, \dots, c^{(K)}) \in \setfnt{C}^K$}
\KwOut{Updated weights $\theta_{\mathrm{out}} \in \setfnt{\Theta}$}
\BlankLine
$(q^{(1)}, \dots, q^{(\kappa)}) \gets (q^{(1)}_{\theta_{\mathrm{in}}}(x), \dots, q^{(\kappa)}_{\theta_{\mathrm{in}}}(x))$\tcp*{Compute query vectors}
$(k^{(1)}, \dots, k^{(K)}) \gets (\psi(c^{(1)}), \dots, \psi(c^{(K)}))$\tcp*{Compute corpus keys (once, then cache + re-use)}
\BlankLine
\tcp{Sample from corpus based on learned notion of relevance:}
\For{$\alpha = 1, \dots, \kappa$}{
   $u^{(\alpha)} \gets \frac{1}{\sqrt{d}}\left( q^{(\alpha)\,\top} k^{(\beta)}:\, \beta=1, \dots, K \right)$\tcp*{Scaled query-key dot products}
   $u^{(\alpha, i^{(1:\alpha-1)})} \gets -\infty$\tcp*{Enforce sampling without replacement}
   $p^{(\alpha)} \gets \softmax(u^{(\alpha)})$\tcp*{Per-item probability of selection}
   $i^{(\alpha)}_{\mathrm{s}} \sim \categorical(p^{(\alpha)})$\tcp*{Stochastically sample corpus index}
   $i^{(\alpha)}_{\mathrm{g}} \gets \argmax_{\beta=1, \dots, K} u^{(\alpha, \beta)}$\tcp*{Greedily select highest-mass index}
   $c^{(\alpha)}_{\mathrm{s}} \gets \operatorname{ToVRAM}(c^{(i^{(\alpha)}_{\mathrm{s}})}),$ $c^{(\alpha)}_{\mathrm{g}} \gets \operatorname{ToVRAM}(c^{(i^{(\alpha)}_{\mathrm{g}})})$\tcp*{Load selected items}
}
$l_{\mathrm{s}} \gets \log \Phi_{\theta_{\mathrm{in}}}\left(y \mid x, c^{(1)}_{\mathrm{s}}, \dots, c^{(\kappa)}_{\mathrm{s}}\right),$ $l_{\mathrm{g}} \gets \log \Phi_{\theta_{\mathrm{in}}}\left(y \mid x, c^{(1)}_{\mathrm{g}}, \dots, c^{(\kappa)}_{\mathrm{g}}\right)$\tcp*{Classifier cross-entropy loss}
$l_{\mathrm{rfc}} \gets -\sum_{\alpha=1}^{\kappa} \log p^{(\alpha, i^{(\alpha)}_{\mathrm{s}})} \operatorname{StopGrad}(l_{\mathrm{s}} - l_{\mathrm{g}})$\tcp*{REINFORCE pseudo-loss}
$g \gets \operatorname{AutoDiff}_{\theta_{\mathrm{in}}}(l_{\mathrm{g}} + l_{\mathrm{rfc}})$\tcp*{Unbiased estimate of $\nabla \mathcal{L}(\theta)$}
$\theta_{\mathrm{out}} \gets \operatorname{OptimizerStep}(\theta_{\mathrm{in}}, g)$\tcp*{Update weights}
\BlankLine
\Return $\theta_{\mathrm{out}}$
\end{algorithm}

In Algorithm \ref{alg:training-step} we provide pseudocode for a single training step of our method. For clarity we omit details of minibatch use and additional implementation details which are described below.

\section{Experimental Details}\label{sec:appendix-experiments}

\subsection{Needle in Haystack}\label{sec:needle-in-haystack}

\begin{figure}[ht]
    \centering
    \includegraphics{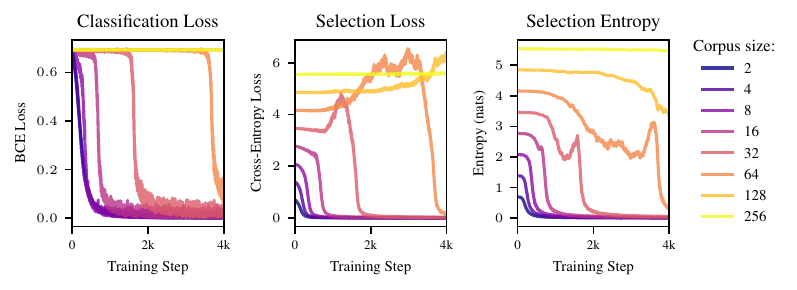}
    \caption{Additional results in the `needle in haystack' setting. (\textbf{left}) Binary cross-entropy classification loss vs. training steps. Loss remains random for some exploration period which scales linearly with the number of historical examples, after which it sharply decreases as the query mechanism learns to retrieve the correct item. (\textbf{center}) Cross-entropy loss of the \emph{retrieval mechanism} over time, when viewing the index of the key-containing historical item as the `true label'. This loss decreases at around the same point at which the classification loss decreases. Note that we do not directly train on this objective -- the retrieval mechanism learns to select the correct item because this leads to lower classification loss. (\textbf{right}) Entropy of the distribution over historical examples from which the retrieval mechanism samples. This sharply increases and then decreases at around the time the classification loss decreases.}
    \label{fig:full-needle-in-haystack}
\end{figure}

Note that the framing of our method in Sec.~\ref{sec:method} can be generalized to cases where the historical data is sampled from a different distribution than $(X_{t_0}, Y_{t_0}).$ Here we aim to construct our setting where our retrieval mechanism must select a particular item from history for the system to achieve nontrivial performance. Towards this end, we relax our setting and allow the sequence of historical items $(X_{t_{-m}}, Y_{t_{-m}}) \to U_{t_{-m}}$ to come from a different distribution.

We generate our dataset in the following manner: we first choose a history size $K \in \Z_{++}.$ We then randomly sample a label $Y_{t_0} \sim \uniform\{0, 1\}$ and a historical example index $m^* \in \uniform\{1, \dots, K\}.$ We let each historical item consist of bitstrings $U_{t_{-m}} = (U_{t_{-m}}^{(1)}, \dots, U_{t_{-m}}^{(D')})$ for $m=1, \dots, K,$ where $U_{t_{-m^*}}^{(1)} = Y_{t_0}$ and $U_{t_{-m}}^{(n)} \sim \uniform\{0, 1\}$ for all $(m, n) \neq (m^*, 1).$ For each historical item we assign a key $\psi(U_{t_{-m}}) \sim \normal(0, 1)^d.$ We let $X_{t_0}$ be a noisy and linearly-transformed version of this key: $X_{t_0} \defeq T \psi(U_{t_{-m}}) + \epsilon$ where $T \sim \normal(0, \tfrac{1}{d})^{d \times d}$ is a fixed linear transform and $\epsilon \sim \normal(0, \sigma_{\mathrm{obs}}^2)^d$ is random noise. Our system can thereby predict $Y$ by predicting $\psi(U_{t_{-m^*}})$ from $X_{t_0},$ retrieving the historical item whose key has maximum cosine similarity with this prediction, then predicting $Y_{t_0}$ equal to its first bit.

In our experiments we sweep $K \in \{2, 4, 8, \dots, 256\},$ and set $D' = 8,$ $d = 64,$ and $\sigma_{\mathrm{obs}} = \sqrt{0.1}.$ We train our models for 4k steps with minibatch size 1k, using the AdamW optimizer with \texttt{lr=2e-4} and other hyperparameters left at their default PyTorch values. Our query generator and classifier are implemented with ReLU MLP blocks, each with a single 512-width hidden layer, and both are fed by a shared 1-layer ReLU MLP input stage. Our system retrieves $\kappa = 1$ historical items.

Additional results are shown in Fig.~\ref{fig:full-needle-in-haystack}. The classification loss is characterized by an exploration phase where the classifier achieves random performance, before its loss sharply decreases. The length of the exploration phase is roughly linear in $K.$ This is consistent with the facts that 1) the classifier can only achieve random performance unless the retrieval mechanism returns the correct item, and 2) the retrieval mechanism has no learning signal unless the classifier has learned to exploit the correct item. Thus, early on learning can only happen when the retrieval mechanism selects the correct item by random chance.

Interestingly, for larger values of $K,$ the selection loss increases during the random exploration phase, suggesting that the mechanism is assigning increasing probability mass to an incorrect subset of historical items. During this time, the selection entropy decreases. At the end of the random exploration phase, the selection entropy sharply increases before decreasing again, suggesting that the retrieval mechanism is unlearning to select the incorrect items before learning to select the correct ones. This phenomenon merits further exploration and likely slows down learning by reducing the rate at which the correct item is retrieved during exploration.

\subsection{Rotating Decision Boundary}\label{sec:nonstationary}

Here we define a binary classification problem where each class is sampled from a distinct Gaussian distribution. The centers of these distributions rotate smoothly around an axis as we vary a time parameter, so that for any given time the optimal decision boundary is a hyperplane, and this hyperplane rotates by $\pi$ radians over the interval of times.

To generate this dataset, we randomly sample an orthonormal pair of vectors $\theta_0 = \theta'_0 / \lVert \theta'_0 \rVert_2$ where $\theta'_0 \sim \normal(0, 1)^D,$ and $\theta_1 = \theta_1' / \lVert \theta_1' \rVert_2$ where $\theta'_1 = \theta''_1 - ((\theta''_1)^\top \theta_0) \theta_0$ and $\theta''_1 \sim \normal(0, 1)^D.$ We then define the rotating class-separation vector $\Delta(t) = \delta (\theta_0 \cos(\pi \tau) + \theta_1 \sin(\pi \tau))$ where $\tau = 0.5 \cos(\pi t) + 0.5,$ for $t \in [0, 1],$ where $\delta > 0.$ For each time $t$ we sample data by first sampling a label $Y_{t_0} \sim \uniform\{0, 1\},$ then covariates $X'_{t_0} = \epsilon + \frac{1}{2} \Delta(t_0) (2Y - 1)$ where $\epsilon \sim \normal(0, \sigma_{\mathrm{obs}})^D.$ We also concatenate a sinusoidal time embedding to the input, as in this setting our system struggles to retrieve appropriate historical items otherwise. Thus, we have $X_{t_0} = (X_{t_0}', \operatorname{embed}(t_0)).$ We let $\psi(X_{t_0}, Y_{t_0}) = X_{t_0}.$

For training samples we let $t_0 \in [0, 0.5],$ and for test samples we let $t_0 \in [0, 1].$ We feed our system sequences of $K$ ordered pairs $(X_{t_{-m}}, Y_{t_{-m}})$ where each $t_{-m} \sim \uniform[0, t_0).$ We set $\delta = 4,$ $\sigma_{\mathrm{obs}} = 1,$ $D = 64,$ 8 Fourier time embedding frequencies, history size $K = 128,$ and let our system retrieve $\kappa = 16$ items. Models are trained for 2500 steps with minibatch size 4096, using the AdamW optimizer with \texttt{lr=5e-5} and other hyperparameters left at their default PyTorch values. Our query generator and classifier are implemented with ReLU MLP blocks, each with 2 512-width hidden layers, and both are fed by a shared 1-layer ReLU MLP input stage. Our `no context' baseline is a ReLU MLP with 3 512-width hidden layers.

\subsection{Amazon Electronics Reviews}

\subsubsection{Dataset}

We use the electronics category of Amazon Reviews '23 \citep{hou2024bridging}, which consists of 43.9M reviews collected from 1998--2023. We sample our training and validation sets from the 4.0M reviews posted before 2014, with test sets sampled from 100 bins uniformly spaced in time from 1998--2023. We use 100k examples as training data which our model learns to predict, and a disjoint set of 1.9M examples as a corpus of historical items to use during training. Our validation set consists of 20k examples which the model trains to predict, and a disjoint 1.9M examples as a corpus of historical items. We generate test data by creating 100 uniformly-spaced time bins and sampling up to 10k examples from each (less for the early time bins which contain $<$10k reviews). For each test set, our historical data comes from a sliding window of the past 1.9M examples. When classifying $X_{t_0},$ we use attention masking to constrain the model to only retrieve historical items from times less than $t_0.$ The historical example corpora are stored in the filesystem as a \texttt{np.memmap} object, with retrieved items loaded into VRAM during the forward pass.

Each review consists of 6 fields: \texttt{title}, \texttt{text}, \texttt{helpful\_vote}, \texttt{verified\_purchase}, and \texttt{timestamp}. We embed the \texttt{title} and \texttt{text} fields to 4096-dimensional vectors using Qwen3-Embedding-8B \citep{zhang2025qwen3}, using the task `Embed the Amazon review for sentiment classification.' and generating the prompt with the Python format string \verb|f'<title>{review["title"]}</title><text>{review["text"]}</text>'|. We use \texttt{helpful\_vote} and \texttt{verified\_purchase} as covariates for our classifier, and generate binary classification labels of 1 (positive sentiment) when \texttt{rating} is above a threshold, and 0 (negative sentiment) when equal or below. While the dataset spans a long duration of time, we do not observe significant performance degradation for classifiers trained on pre-2014 data and evaluated on our test sets, possibly due to the text embedding model being trained on data beyond 2023. Thus, to increase the distribution shift, we set our positive sentiment threshold to 3 for pre-2014 reviews and 1 for post-2014 reviews.

\subsubsection{Architecture and Implementation Details}

\begin{figure}[ht]
   \centering
   \includegraphics{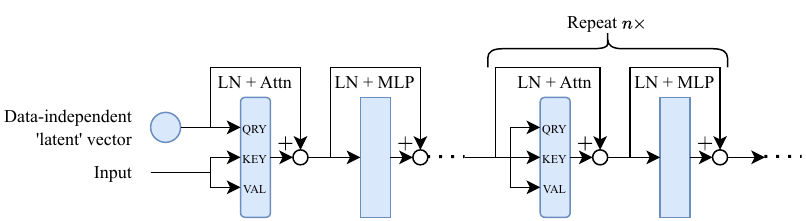}
   \caption{Diagram of a Perceiver block \citep{jaegle2021perceiver}. Rather than attending directly to the full input sequence like a standard transformer, Perceivers use cross-attention to combine the input with a learned data-independent `latent' vector. They then apply a sequence of standard transformer blocks to this sequence. This approach reduces the computational cost from $L^2$ to $LM$ where $L$ is the input sequence length and $M$ is the latent sequence length (a hyperparameter). The Perceiver architecture is appealing for our use case because it was proposed with multimodal feature fusion in mind, and because it is more-scalable to long sequences than standard transformers.}
   \label{fig:perceiver-block}
\end{figure}

We implement the query generator and classifier of our system using Perceiver blocks \citep{jaegle2021perceiver} (illustrated in Fig.~\ref{fig:perceiver-block}). We prefer these over standard transformer blocks because they were proposed with multimodal data in mind, and have better scalability to long input sequences. However, our method is architecture-agnostic. We use an embedding dimension of $512,$ a latent sequence length of $128,$ and stages consisting of 1 cross-attention block followed by 2 self-attention blocks. We retrieve $\kappa = 4$ historical items.

Corpus keys are generated by taking the first 64 dimensions of the embedded text features. Note that the Qwen3-Embedding models use Matryoshka representation learning \citep{kusupati2022matryoshka}, which is designed to facilitate dimensionality reduction through truncation in this manner.

We preprocess text embeddings by standardizing them, then linearly projecting them to the embedding dimension of our Perceiver blocks. For the \texttt{helpful\_votes} feature, we standardize it, apply a Fourier embedding with 48 frequency bands and a maximum frequency of 64, then linearly project it to the embedding dimension. For the \texttt{verified\_purchase} feature, we simply linearly project it to the embedding dimension.

We apply the following bag of tricks, which were found helpful in preliminary experiments on similar datasets:
\begin{itemize}
    \item We find that pure cosine similarity between the embeddings of the text feature of $X_{t_0}$ and those of the historical items performs quite well. Thus, instead of training our retrieval mechanism from scratch, we train it to learn the residual with this cosine similarity: we parameterize it as $q_\theta(x) = \alpha \tilde{q}_\theta(x) + (1-\alpha) q_{\mathrm{sim}}(x),$ where $\alpha \in (0, 1)$ is learned and $q_{\mathrm{sim}}$ simply returns the first 64 dimensions of the text embedding.
    \item We randomly drop out the direct input $\to$ classifier connection with probability \texttt{cls\_stage\_dropout\_rate = 0.9} (similarly to stochastic depth \citep{huang2016deep}), forcing the model to rely solely on the retrieved items for classification. This is helpful because it counteracts a tendency for the classifier to learn to ignore retrieved items early in training when they are not relevant, thereby depriving the retrieval mechanism of a learning signal.
    \item We randomly drop out some of the retrieved items with rate \texttt{query\_dropout\_rate = 0.1}.
    \item We use a $10\times$ higher learning rate for the parameters of the retrieval mechanism than for the rest of the system.
    \item Over the course of training we exponentially decay the temperature of the retrieval mechanism's softmax distribution from a starting temperature of \texttt{max\_temperature = 0.01} to a final temperature of \texttt{min\_temperature = 0.001}. This is essentially a form of curriculum learning, where early in training we have low-variance gradients which are easy to train with, and late in training we have higher-variance but retrieval mechanism behavior closer to the zero-temperature limit used at test time.
    \item We use AdamW-style weight decay with strength $10^{-4},$ apply gradient norm clipping with maximum norm $1,$ and a linear learning rate warmup for the first 10\% of epochs, followed by cosine decay down to $0.1\times$ the base learning rate for the remaining epochs.
\end{itemize}
We also explored the following tricks, but found them unhelpful in this setting:
\begin{itemize}
    \item In order to prevent premature collapse of the retrieval mechanism to a bad solution, we add a term proportional to the negative entropy of the retrieval mechanism's softmax distribution to the loss, with strength which decreases over the course of training.
    \item In order to encourage the $\kappa$ queries to retrieve diverse historical examples rather than selecting based on similar criteria, we add a term proportional to the mean probability mass assigned by a given query to the items selected by different queries.
    \item In order to bias the retrieval mechanism towards selecting based on the cosine similarity between the first 64 dimensions of the text embeddings of $X_{t_0}$ and those of the historical items, we add a term proportional to the cosine similarity between our query vectors and the first 64 text embedding dimensions of $X_{t_0}.$
\end{itemize}
We train our system for 100 epochs with minibatch size 1024 using the AdamW optimizer with base learning rate \texttt{lr=1e-3}.

\subsubsection{Impact of Retrieving Full Historical Examples}\label{sec:label-vs-nolabel}

\begin{figure}[ht]
    \centering

    \makebox[\textwidth]{
        \makebox[0.5\textwidth][c]{\hspace{2em}Full retrieval of $(X_{t_{-m}}, Y_{t_{-m}})$}
        \makebox[0.5\textwidth][c]{\hspace{5em}Label-only retrieval of $Y_{t_{-m}}$}
    }
    \vspace{-1em}
    
    \includegraphics{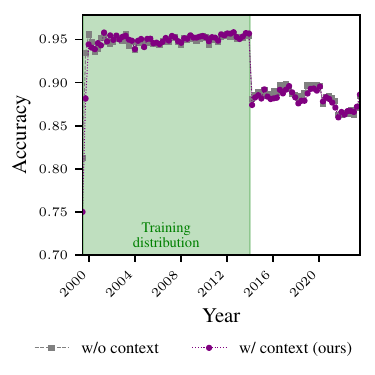}
    \hfill
    \includegraphics{figures/context_vs_nocontext_amazon_reviews.pdf}
    \caption{Accuracy over time for models trained on pre-2014 data from the electronics category of Amazon Reviews '23. (\textbf{left}) Our retrieval mechanism returns full historical examples $(X_{t_{-m}}, Y_{t_{-m}}).$ (\textbf{right}) Our retrieval mechanism returns only the labels $Y_{t_{-m}}$ of historical examples. In the former case our system has similar robustness to a standard classifier, while in the latter case it is able to leverage historical context to mitigate performance degradation on the out-of-distribution data beyond 2014.}
    \label{fig:amazon-reviews-label}
\end{figure}

As shown in Fig.~\ref{fig:amazon-reviews-label}, retrieving full historical examples $(X_{t_{-m}}, Y_{t_{-m}})$ does not improve robustness over the baseline. However, retrieving only the labels $Y_{t_{-m}}$ does improve performance. This suggests that in this particular setting, the full features of historical examples have little utility for classifying $X_{t_0},$ and drown out the useful signal provided by the historical labels with noise or alternate non-generalizing signals. This indicates the need for better regularization or architectural strategies to enable the classifier to leverage both historical features and labels.

\section{Derivation of Score-Based Gradient Estimator and Application to Selection Mechanism}\label{sec:reinforce-derivation}

The derivations below are not novel, but for completeness and ease of reading we include them here in our own notation. For brevity we omit engineering details such as the baseline used to reduce the variance of the gradient estimator.

Note that this gradient estimator is sometimes referred to as `REINFORCE', and is a building block for many reinforcement learning algorithms. Our setup can be viewed as episodic reinforcement learning with episode length 1, the action being the selection of items from the corpus, and the reward given by the negative loss of the downstream classifier. 

\subsection{Gradient Estimation in the Normal Deep Learning Context}

For context, we will first discuss gradient estimation in the normal supervised deep learning setting. Suppose we have a space of inputs $\setfnt{X} \subset \R^D$ and outputs $\setfnt{Y} \subset \R^M.$ We view inputs and outputs as jointly-distributed random variables $X \in \setfnt{X}$ and $Y \in \setfnt{Y}$ with $(X, Y) \sim p$ for some data-generating distribution $p.$ We aim to train a classifier $\Phi_\theta: \setfnt{X} \to \hat{\setfnt{Y}}$ parameterized by weights $\theta \in \setfnt{\Theta} \subset \R^P,$ where $\hat{\setfnt{Y}} \subset \R^{M'}$ is a space of predicted outputs which can be mapped to outputs in $\setfnt{Y}.$ The classifier is trained to minimize some loss function $\ell: \setfnt{\Theta} \times \setfnt{X} \times \setfnt{Y} \to \R$ in expectation over the data distribution:
\begin{equation}
   \min_{\theta \in \setfnt{\Theta}} \quad \mathcal{L}(\theta) \defeq \expec_{X, Y \sim p} \ell(\theta; X, Y).
\end{equation}
For example, when training an ImageNet classifier, $\setfnt{X} \subset \R^{3 \cdot 224 \cdot 224}$ might represent the space of images, $\setfnt{Y} \equiv \{1, \dots, 1000\}$ the space of labels, $\hat{\setfnt{Y}} \equiv \Delta^{999}$ the probability simplex over 1000 classes, and $\ell(\theta; x, y) \defeq -\log \Phi_\theta(y \mid x)$ the cross-entropy loss of the classifier.

We approximately solve this optimization problem using stochastic gradient descent. Note that because $p$ does not depend on $\theta,$ we can exchange the order of differentiation and expectation:
\begin{equation}\label{eqn:expectation-differentiation-exchange}
   \nabla \mathcal{L}(\theta) = \nabla_\theta \expec_{X, Y \sim p} \ell(\theta; X, Y) = \expec_{X, Y \sim p} \nabla_\theta \ell(\theta; X, Y).
\end{equation}
While analytically integrating over the data distribution $p$ is intractable, we can use a finite number of datapoints $(x^{(1)}, y^{(1)}), \dots, (x^{(N)}, y^{(N)}) \overset{\mathrm{i.i.d.}}{\sim} p$ to approximate the expectation:
\begin{equation}
   \nabla \mathcal{L}(\theta) = \expec_{X, Y \sim p} \nabla_\theta \ell(\theta; X, Y) \approx \frac{1}{N} \sum_{\alpha=1}^{N} \nabla_\theta \ell(\theta; x^{(\alpha)}, y^{(\alpha)}).
\end{equation}
Each term in the summation is now straightforward to compute with automatic differentiation \citep{paszke2019pytorch}.

\subsection{Gradient Estimation when our Model contains a Discrete Stochastic Layer}

Now suppose $\Phi_\theta$ contains a discrete stochastic layer. We will denote by $\phi_\theta: \setfnt{X} \to \Delta^{N - 1}$ a deterministic function in our model which generates an $X$-dependent random variable $U$ such that $U \mid X \sim p_\theta(\cdot \mid X) \defeq \categorical(\phi_\theta(X)),$ and $\psi_\theta: \setfnt{X} \times \{1, \dots, N\} \to \hat{\setfnt{Y}}$ a downstream function which predicts the output based on $X$ and a sample from $p_\theta(\cdot \mid X).$ Now $\Phi_\theta(X)$ is a random variable $\psi_\theta(X, U)$ where $U \sim p_\theta(\cdot \mid X).$ We let our loss function depend on the sample from our stochastic layer, $\ell: \setfnt{\Theta} \times \setfnt{X} \times \setfnt{Y} \times \{1, \dots, N\} \to \R,$ and our objective now contains an expectation over both the data distribution and the randomness of our model:
\begin{equation}
   \mathcal{L}(\theta) \defeq \expec_{X, Y \sim p,\, U \sim p_\theta(\cdot \mid X)} \ell(\theta; X, Y, U).
\end{equation}
In the ImageNet example, we would have $\ell(\theta; x, y, u) = -\log \psi_\theta(y \mid x, u).$

Here both expectations are intractable or impractical to compute analytically. However, we cannot simply exchange the order of differentiation and expectation with respect to the second integral the way we did above, i.e.
\begin{align}
   \nabla \mathcal{L}(\theta) &= \nabla_\theta \expec_{X, Y \sim p,\, U \sim p_\theta(\cdot \mid X)} \ell(\theta; X, Y, U) &\\
   &= \expec_{X, Y \sim p} \nabla_\theta \expec_{U \sim p_\theta(\cdot \mid X)} \ell(\theta; X, Y, U) \\
   &\neq \expec_{X, Y \sim p,\, U \sim p_\theta(\cdot \mid X)} \nabla_\theta \ell(\theta; X, Y, U),
\end{align}
because $p_\theta(\cdot \mid X)$ depends on $\theta.$ Instead, we use the REINFORCE trick:
\begin{align}
   & \nabla_\theta \expec_{U \sim p_\theta(\cdot \mid X)} \ell(\theta; X, Y, U) &\\
   = \quad & \nabla_\theta \sum_{u = 1}^{N} p_\theta(u \mid X) \ell(\theta; X, Y, u) \\
   = \quad & \sum_{u = 1}^{N} \Big[ \nabla_\theta p_\theta(u \mid X) \ell(\theta; X, Y, u) + p_\theta(u \mid X) \nabla_\theta \ell(\theta; X, Y, u) \Big] \\
   = \quad & \sum_{u = 1}^{N} \Big[ p_\theta(u \mid X) \nabla_\theta \log p_\theta(u \mid X) \ell(\theta; X, Y, u) + p_\theta(u \mid X) \nabla_\theta \ell(\theta; X, Y, u) \Big] \\
   = \quad & \expec_{U \sim p_\theta(\cdot \mid U)} \Big[ \nabla_\theta \log p_\theta(u \mid X) \ell(\theta; X, Y, u) + \nabla_\theta \ell(\theta; X, Y, u) \Big].
\end{align}
This expression allows us to approximate the expectation with a finite number of datapoints $(x^{(1)}, y^{(1)}), \dots, (x^{(N)}, y^{(N)}) \overset{\mathrm{i.i.d.}}{\sim} p$ and samples from our stochastic layer $u^{(1)} \sim p_\theta(\cdot \mid x^{(1)}), \dots, u^{(N)} \sim p_\theta(\cdot \mid x^{(N)}),$ similarly to above:
\begin{align}
   \nabla \mathcal{L}(\theta) &= \expec_{X, Y \sim p,\, U \sim p_\theta(\cdot \mid X)} \Big[ \nabla_\theta \log p_\theta(U \mid X) \ell(\theta; X, Y, U) + \nabla_\theta \ell(\theta; X, Y, U) \Big] &\\
   &\approx \frac{1}{N} \sum_{\alpha = 1}^{N} \Big[ \nabla_\theta \log p_\theta(u^{(\alpha)} \mid x^{(\alpha)}) \ell(\theta; x^{(\alpha)}, y^{(\alpha)}, u^{(\alpha)}) + \nabla_\theta \ell(\theta; x^{(\alpha)}, y^{(\alpha)}, u^{(\alpha)}) \Big].\label{eqn:reinforce-autodiff-amenable}
\end{align}
We can compute the quantities $\nabla_\theta \log p_\theta(u \mid X) = \nabla_\theta \log \phi_\theta(u \mid X)$ with automatic differentiation; thus, Eqn.~\ref{eqn:reinforce-autodiff-amenable} enables us to train our model with stochastic gradient descent. In PyTorch, this can be implemented by backpropagating through the pseudo-loss function
\begin{align}
   \ell_{\mathrm{pseudo}}(\theta; x, y, u) &= \log p_\theta(u \mid x) \operatorname{StopGrad}(\ell(\theta; x, y, u)) + \ell(\theta; x, y, u) &\\
   \text{where} \quad u &= \operatorname{StopGrad}(\tilde{u}),\, \tilde{u} \sim p_\theta(\cdot \mid x).
\end{align}

\subsection{Application to our Retrieval Mechanism}

Our method fits into the above framework as follows: we let the tuple $X \gets (X_{t_0}, X_{t_{-1}}, Y_{t_{-1}}, X_{t_{-2}}, Y_{t_{-2}}, \dots, X_{t_{-K}}, Y_{t_{-K}})$ be the input and $Y \gets Y_{t_0}$ be the output. Our stochastic layer is represented by the retrieval mechanism, which we will denote $p_\theta(X),$ which gives the distribution over the historical example indices $U \gets \vecfnt{I} \equiv (I^{(1)}, \dots, I^{(\kappa)})$ where each $I^{(m)} \in \{1, \dots, K\}.$ Note that $\vecfnt{I}$ fits into the above framing because it is a bijection of a categorical random scalar over $N \equiv \prod_{m = 1}^{\kappa} (K - m + 1)$ categories. The downstream function is then given by $\psi_\theta(X, U) \gets \Phi_\theta(Y_{t_0} \mid X_{t_0}, X_{\hat{t}_{1}}, Y_{\hat{t}_{1}}, \dots, X_{\hat{t}_{\kappa}}, Y_{\hat{t}_{\kappa}}).$

Thus, by sampling a finite number of inputs and historical sequences $(x_{t_0}, y_{t_0}, x_{t_{-1}}, y_{t_{-1}}, \dots, x_{t_{-K}}, y_{t_{-K}})$ from the data distribution, we can perform stochastic gradient descent by leveraging Eqn.~\ref{eqn:reinforce-autodiff-amenable}. 

\end{document}